\pdfoutput=1

\documentclass[11pt]{article}

\usepackage{EMNLP2023}

\usepackage{times}
\usepackage{latexsym}
\usepackage{float}

\usepackage[utf8]{inputenc}
\usepackage{tikz}
\usetikzlibrary{trees}

\usepackage{tikz}
\usetikzlibrary{positioning,shapes.geometric}

\usepackage[T1]{fontenc}

\usepackage[utf8]{inputenc}

\usepackage{microtype}

\usepackage{inconsolata}
\usepackage{graphicx}
\usepackage{booktabs}
\usepackage{amssymb} 
\usepackage{tabularx} 

\usepackage{rotating}
\usepackage{hyperref}

\usepackage{xcolor}
\usepackage[normalem]{ulem}

\def\MEdel#1{\bgroup\markoverwith{\textcolor{red}{\rule[0.5ex]{2pt}{1pt}}}\ULon{#1}}

\title{Arcee's MergeKit: A Toolkit for Merging Large Language Models}

\author{Charles Goddard,  Shamane Siriwardhana, Malikeh Ehghaghi, Luke Meyers, \\
{\bf Vlad Karpukhin, Brian Benedict, Mark McQuade, Jacob Solawetz  } \\
Arcee, Florida, USA \\
\texttt{\{charles, shamane, malikeh, luke, vlad, benedict, mark, jacob\}@arcee.ai}}

\begin{document}
\maketitle
\begin{abstract}

The rapid growth of open-source language models provides the opportunity to merge model checkpoints, combining their parameters to improve performance and versatility. Advances in transfer learning have led to numerous task-specific models, which model merging can integrate into powerful multitask models without additional training. MergeKit is an open-source library designed to support this process with an efficient and extensible framework suitable for any hardware. It has facilitated the merging of thousands of models, contributing to some of the world's most powerful open-source model checkpoints. The library is accessible at: \href{https://github.com/arcee-ai/mergekit}{https://github.com/arcee-ai/mergekit}.
\end{abstract}

\section{Introduction}
Over the past year, open-source Large Language Models (LLMs) have rapidly developed and are accessible via the Hugging Face model hub \cite{wolf2019huggingface}. These models, trained on up to trillions of tokens, typically range from 1-70+ billion parameters \cite{minaee2024large, zhang2024tinyllama}. Open-source checkpoints include pretrained and instruction-tuned models across domains like coding \cite{roziere2023code} and medical applications \cite{wu2023pmc}. Fine-tuning separate models for each task presents challenges: storing and deploying each model separately and the inability of independently trained models to leverage insights from related tasks \cite{sanh2021multitask, rame2023model, yadav2024ties, yu2023language}.

Training these models from scratch requires substantial investment. Further fine-tuning can lead to catastrophic forgetting \cite{de2021continual}, degrading their general capabilities and performances across tasks \cite{cheng2023adapting, wu2024llama}. Aligning models to respond favorably requires extensive human preference data, often unattainable for most teams \cite{wang2023aligning, rafailov2024direct}. This raises the question of leveraging existing pretrained checkpoints. Model merging has emerged as a transformative strategy, combining parameters from multiple models into a single one, enabling multitask and continual learning while reducing catastrophic forgetting \cite{siriwardhana2024domain}.

In this paper, we introduce MergeKit\footnote{\href{https://github.com/arcee-ai/mergekit}{https://github.com/arcee-ai/mergekit}}, a centralized library for executing community-formulated merging strategies, compatible with memory-constrained CPUs and accelerated GPUs. Our main contributions are: (1) an overview of current model merging research to date and (2) a presentation of MergeKit's key objectives, architectural decisions, and development principles to establish an extensible foundation for the future efforts of the model merging community.








\section{Background \& Related Work}

\subsection{The Concept of Model Merging}

Model merging \cite{ainsworth2022git}, a recent focus in research, integrates two or more pretrained models into a unified model that retains their strengths. This concept builds on weight averaging \cite{utans1996weight} and mode connectivity \cite{garipov2018loss}. Techniques often leverage Linear Mode Connectivity (LMC) \cite{entezari2021role} for models fine-tuned from a common pretrained model \cite{nagarajan2019uniform, neyshabur2021transferred}. Other works employ permutation equivariance and apply transformations to model weights, aligning them in the loss landscape \cite{ainsworth2022git, stoica2023zipit, verma2024merging}.

\subsection{Different Types of Model Merging}

\begin{figure*}[ht]
\centering
\vspace{-15pt}
\includegraphics[width=1.0\textwidth]{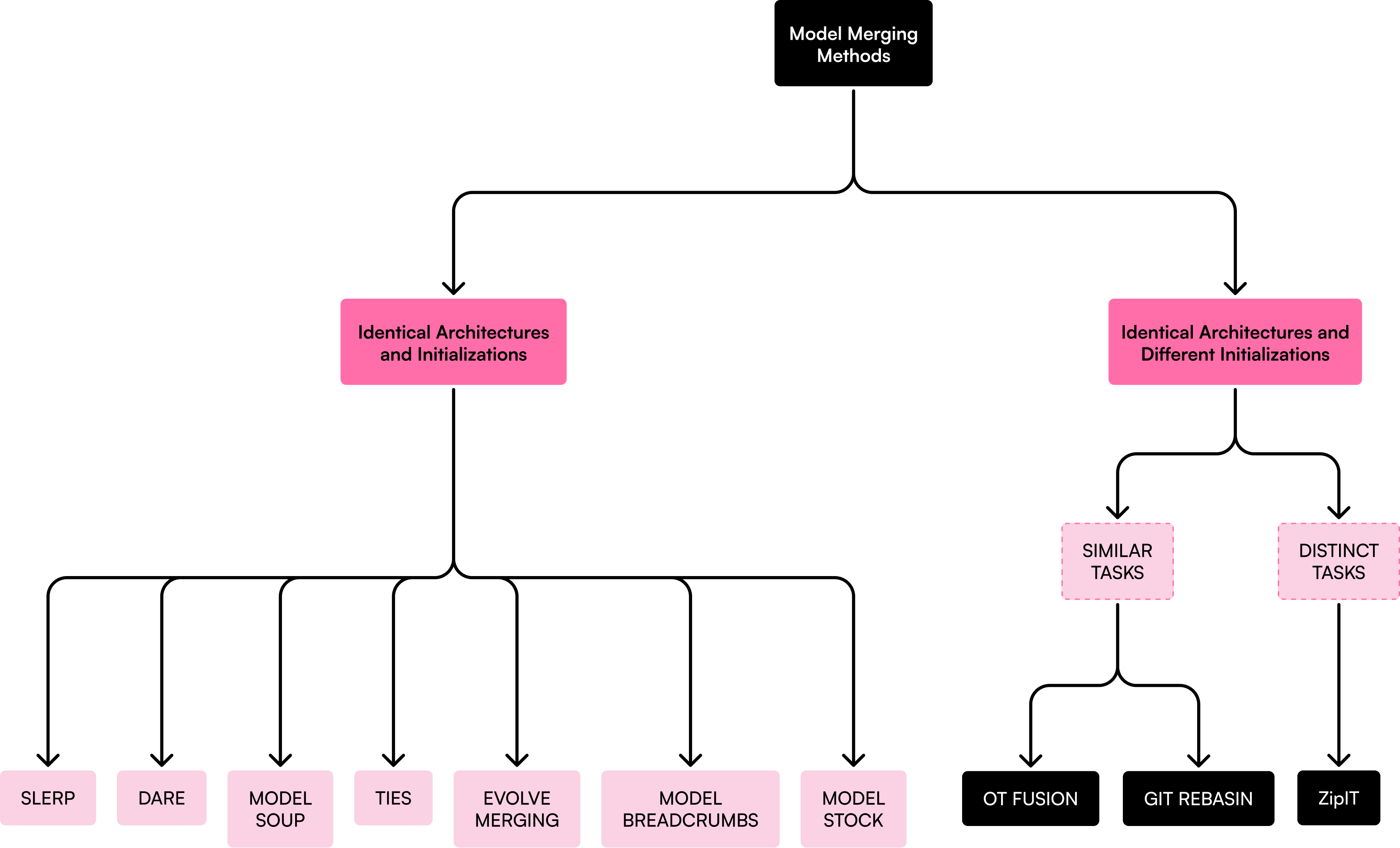}
\caption{Classification of model merging methods. We currently support the model merging methods outlined on the left, and we are actively working to incorporate additional merging techniques such as ZipIt, OT Fusion, and Git Rebasin.}
\label{fig:Merge_Classification}
\vspace{-15pt}
\end{figure*}

In developing our toolkit, as shown in Figure \ref{fig:Merge_Classification}, we categorize existing and anticipated model merging techniques. This classification enhances understanding by focusing on two critical aspects: weight initializations and the architectural configurations of various checkpoints.

\subsubsection{Merging Models with Both Identical Architectures and Initializations}

This section explores model merging techniques using LMC \cite{nagarajan2019uniform} to derive a final merged model through linear interpolation. A key requirement is that the models must have identical architectures and initializations.

The simplest method, built upon the results of weight averaging literature \cite{utans1996weight, smith2017investigation, garipov2018loss, izmailov2018averaging} and the Model Soups \cite{wortsman2022model} approach, is linear averaging of weights. This technique relies on linear mode connectivity and is the foundation of most others. 

Task Arithmetic \cite{ilharco2022editing} expands upon this approach by introducing the concept of task vectors, showing that performing arithmetic on the differences between fine-tuned models and a common base model is both useful and semantically meaningful. 

Trim, Elect Sign \& Merge (TIES merging) \cite{yadav2023resolving}, Model Breadcrumbs \cite{davari2023model}, and Drop And
REscale (DARE) \cite{yu2023language} further introduce methods for sparsifying and combining these task vectors that enable larger numbers of models to be combined into one without degrading capabilities.

The use of the Spherical Linear intERPolation (SLERP) technique \cite{shoemake1985animating} to interpolate between model checkpoints is an extension of simple weight averaging. Its success shows that there is often a spherical path with a lower loss barrier than a direct linear interpolation. SLERP\footnote{\href{https://github.com/Digitous/LLM-SLERP-Merge}{https://github.com/Digitous/LLM-SLERP-Merge}} leverages the geometric and rotational properties within the models' vector space, ensuring a blend that more accurately embodies the characteristics of both parent models.

Other approaches introduce weighting factors defined in terms of model activations that must be computed with training data. \citet{matena2022merging} explore the use of the Fisher information matrix. \citet{jin2022dataless} introduce the Regression Mean (RegMean) method, which allows merges to produce optimal weights with respect to $L2$ distance to model predictions while keeping training data private.


MergeKit introduces two novel methods for building larger models without performing any parameter-space combination. Referred to online as `FrankenMerging', the passthrough method in MergeKit allows the piecewise combination of layers from multiple models into a new model of unusual size. This technique is behind the popular model Goliath-120b\footnote{\href{https://huggingface.co/alpindale/goliath-120b}{alpindale/goliath-120b}}, and is the first step of the Depth Up-Scaling technique of \cite{kim2023solar} used for SOLAR-10.7B\footnote{\href{https://huggingface.co/upstage/SOLAR-10.7B-v1.0}{upstage/SOLAR-10.7B-v1.0}} and Yi-9B\footnote{\href{https://huggingface.co/01-ai/Yi-1.5-9B}{01-ai/Yi-1.5-9B}}. Similarly referred to as Franken Mixture of Experts (`FrankenMoE'), the \texttt{mergekit-moe} script allows building a Mixture of Experts (MoE) model from multiple dense models using either a prompt based hidden state heuristic for semantic routing or randomly initialized gates for sparse up-cycling as in \cite{komatsuzaki2023sparseupcyclingtrainingmixtureofexperts}.

Evolutionary Model Merging~\cite{akiba2024evolutionary} is a novel method that automates the creation of foundation models by leveraging diverse open-source models without extensive additional training data. This approach optimizes combining models from different domains in both parameter space (PS) and data flow space (DFS). PS optimization integrates the weights of multiple models, while DFS preserves original weights and optimizes the inference path. Models created using evolutionary model merging, such as EvoLLM-JP~\cite{akiba2024evolutionary}, demonstrate state-of-the-art performance, highlighting the efficiency and generalizability of this technique.


\subsubsection{Merging Models with Identical Architectures and Different Initializations}

This section explores advanced merging methods beyond combining checkpoints with identical initializations. Previous research shows that simple linear model combination is insufficient for different initializations~\cite{ainsworth2022git}. Methods leveraging permutation symmetry of checkpoints include Git-Rebasin~\cite{ainsworth2022git} and Optimizing Mode Connectivity via Neuron Alignment~\cite{tatro2020optimizing}, which permute weights of independently trained models to reduce interpolation barriers. Optimal Transport Fusion (OTFusion)~\cite{singh2020model} operates similarly but computes a soft mapping between neurons using Optimal Transport. These methods assign correspondences between model neurons and perform simple interpolation in transformed weight space. Recent work~\cite{imfeld2023transformer, verma2024merging} extends these methods to Transformer-based models. \cite{jordan2022repair} addresses variance collapse in interpolated networks with a rescaling step, reducing loss barriers between permuted models. ZipIt~\cite{stoica2023zipit} expands the scope by merging models with similar architectures trained on distinct tasks. This method correlates features within and across models, and can also allow partial merging to create a multi-head model. ZipIt preserves and integrates knowledge from different domains into a unified model without additional training.

These techniques do not yet share the wide adoption and success of merging models trained from a common initialization, but present a promising future research direction for the field of merging.

\subsubsection{Fusing Models with Different Architectures}

While not strictly model merging, Composition to Augment Language Models (CALM) \cite{bansal2024llm} and knowledge fusion approaches like FUSELLM \cite{wan2024knowledge} advance the fusion of models with diverse architectures. CALM uses cross-attention mechanisms to blend representations from different models, leveraging their combined strengths across varied neural network structures. FUSELLM focuses on aligning and fusing the probabilistic distributions of source LLMs to amplify their collective knowledge and advantages. Unlike previous methods, these approaches require additional training of the models.

\subsection{Practical Use Cases of Model Merging}

Model merging significantly impacts machine learning models on platforms like Hugging Face \cite{wolf2019huggingface}. Merged models, such as BioMistral \cite{labrak2024biomistral}, Aloe \cite{gururajan2024aloe}, Llama-3-SEC~\cite{siriwardhana2024domain}, Prometheus 2 \cite{kim2024prometheus}, and OpenPipe's Mistral 7B Fine-Tune Optimized \cite{Corbitt_2023}, demonstrate competitive performance in specialized domains and fine-tuning applications. \citet{wei2024opdai} highlight merging's success in enhancing hallucination detection performance. \citet{tao2024unlocking} show effectiveness of model merging to develop task-solving LLMs for low-resource languages. The success of merged models underscores their value in continuous and multitask learning, enabling the creation of versatile models that excel at multiple tasks or adapt to new domains without retraining from scratch. This approach maximizes existing resources and fosters innovative solutions for complex problems.

\section{Library Design: Key Design Principles}

MergeKit has been thoughtfully engineered to facilitate the straightforward application of both current and forthcoming model merging techniques. Our repository includes detailed tutorials and IPython\footnote{\href{https://github.com/arcee-ai/mergekit/blob/main/notebook.ipynb}{https://github.com/arcee-ai/mergekit/blob/main/notebook.ipynb}} notebooks to guide users through the process of utilizing MergeKit effectively. This section is dedicated to outlining the fundamental design principles underpinning the library, with the aim of assisting the open-source community in adopting our toolkit and incorporating new techniques.

\begin{figure*}[ht]
\centering
\includegraphics[width=0.8\textwidth]{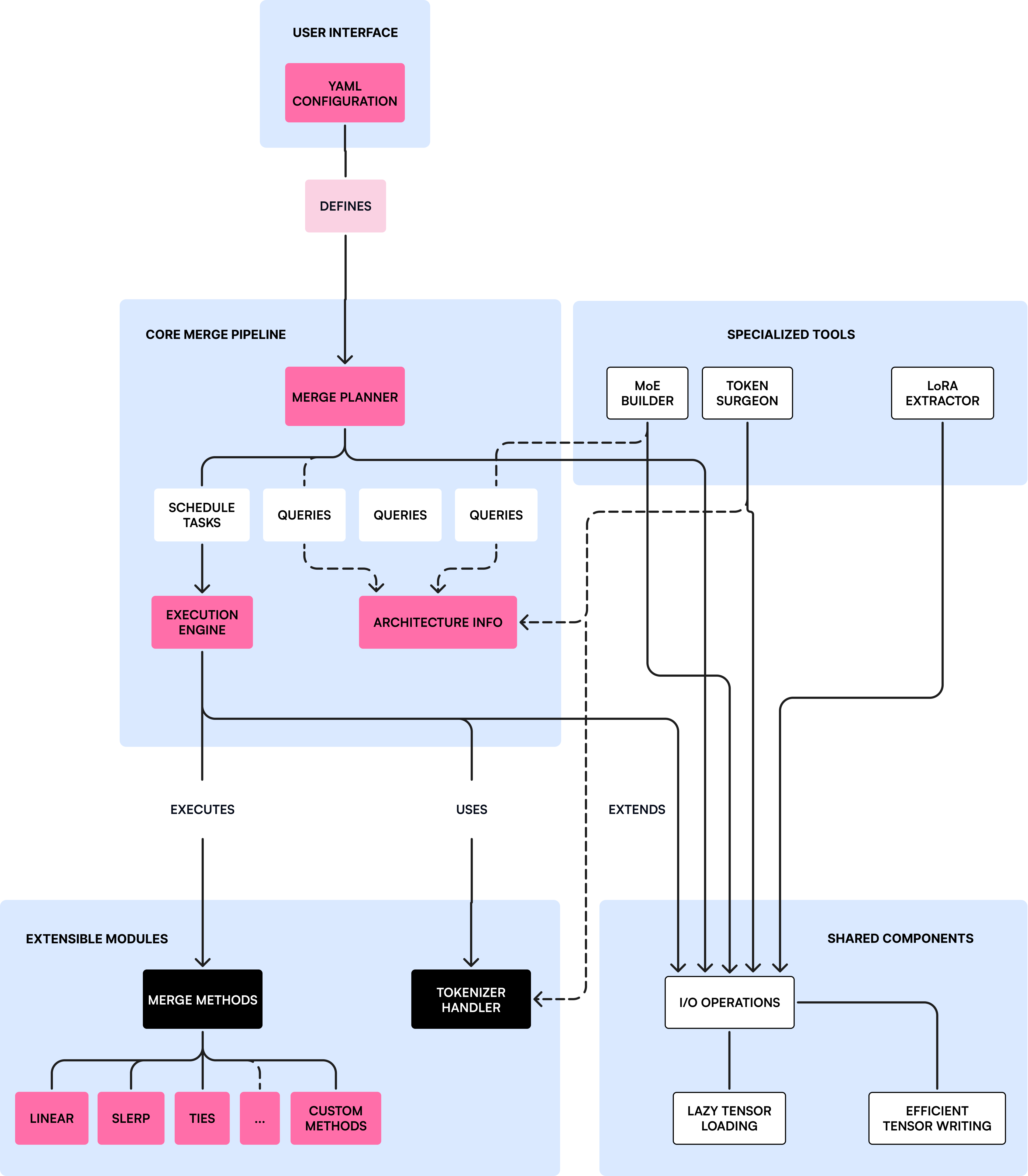}
\caption{MergeKit Architecture. \small{\raggedright The diagram depicts the software architecture of MergeKit and highlights the points meant to be extended and components that are easily reusable in other scripts.}}

\label{fig:MergeKit-repo-structure}
\end{figure*}

\subsection{User-Centric Design: Intuitive Interface and YAML Configuration Control}
The primary interface for MergeKit is through YAML configuration files that allow users of all skill levels to define complex merge operations without the need for coding experience. This approach both democratizes the use of MergeKit and fosters community engagement by making merge recipes easily repeatable, shareable, and remixable.

A YAML\footnote{\href{https://github.com/arcee-ai/mergekit/tree/main/examples}{https://github.com/arcee-ai/mergekit/tree/main/examples}} configuration file defines the merge method, input models, and any parameters necessary for the merging algorithm selected. Parameters can be set globally or targeted to specific model components, and can be specified as constant scalar values or as layer-varying interpolated gradients. These different levels of granularity offer an easy introduction for simple merges while allowing power users to define truly complex operations.

\subsection{Modularity: Plug-and-Play Components}
MergeKit is designed with composability and reusability as guiding principles. Merge methods are designed to be interchangeable and easy-to-add. Components are structured such that they can be added, removed, or interchanged to allow customization and experimentation. Wherever possible, components are designed to be useful standalone for external use. For instance, MergeKit's lazy tensor loading functionality is a core component of the toolkit, but is also simple and convenient to pull into one-off scripts. Figure \ref{fig:MergeKit-repo-structure} highlights some important points of extensibility and reusable components. MergeKit is tightly integrated with the Hugging Face Transformers library \cite{wolf2019huggingface} and its model hub.


\subsection{Scalability: Efficiency and Performance Optimization}

MergeKit is designed specifically to address the challenge of merging large pretrained language models with a diverse range of available computational resources. At the heart of its efficiency is an out-of-core approach to model merging. By loading only the tensors necessary for each individual operation into working memory, MergeKit can scale from a high-end research cluster all the way down to a personal laptop with no GPU and limited Random-Access Memory (RAM).  We use Directed Acyclic Graph (DAG) approach to optimize the merging process for large models. The DAG structure allows for efficient computation by organizing operations in a way that minimizes redundancy and resource usage. This method is particularly advantageous in handling model merging on resource-constrained environments.

\subsubsection{Computational Graph Scheduling}

MergeKit internally represents a merge as a directed acyclic graph of operations, or Task instances. This representation is used to schedule the execution of tasks such that the working set needed at any given time is minimized. Execution of the graph also implicitly handles eviction of intermediate values that are no longer needed. This infrastructure allows developers to build new merge methods that benefit from MergeKit's memory efficiency and hardware scalability with little to no extra effort.


\subsection{Mergekit Graphical User Interface (GUI)}

\begin{figure*}[ht]
\centering
\vspace{-15pt}
\includegraphics[width=1.0\textwidth]{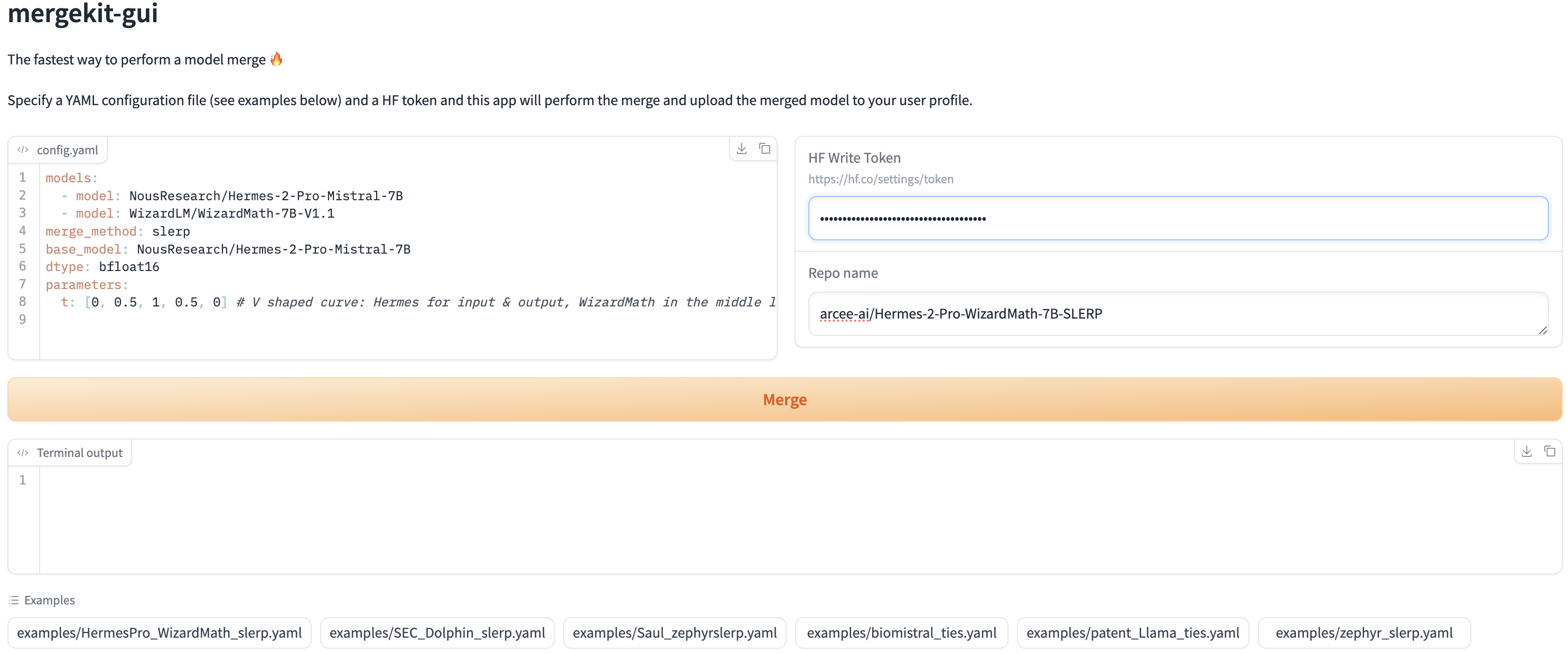}
\caption{Demo of MergeKit-GUI.}
\label{fig:mergekit_demo}
\vspace{-15pt}
\end{figure*}

We developed MergeKit-GUI\footnote{\href{https://huggingface.co/spaces/arcee-ai/mergekit-gui}{https://huggingface.co/spaces/arcee-ai/mergekit-gui}}, a user-friendly interface hosted on Hugging Face running on A100 GPU, designed to simplify the model merging process. With this GUI, users can easily upload configuration files, select from an array of different merging techniques, and execute merges with a few clicks. A demonstration of MergeKit-GUI is shown in Figure \ref{fig:mergekit_demo}.  

The workflow is straightforward: users start by uploading a YAML configuration file—either by providing their own or by choosing from a variety of pre-configured examples available on the interface. After the configuration file is set, users input their Hugging Face token for authentication and specify the repository name where the final merged model will be stored.

Once all parameters are configured, users can click on the `Merge' button to initiate the process. The terminal output displays real-time logs, allowing users to monitor the merging process step-by-step. Upon successful completion, the following confirmation message appears: \\
\texttt{Process completed successfully. Model successfully uploaded to HF: <REPOSITORY\_NAME>}.

\section{Extensibility of MergeKit}

Given the rapid success of model merging techniques and the anticipated development of innovative methods, we invite the community to develop novel merging strategies and enhancements, thereby contributing to the growth and refinement of MergeKit. This section aims to provide a streamlined guide on integrating new merging methods into MergeKit, utilizing existing functionalities where applicable to facilitate the process.

To incorporate a new merging method into MergeKit, contributors should familiarize themselves with several key Python modules within the repository:

\begin{itemize}

  \item \texttt{merge\_methods/base.py}: Defines the interface that new merge methods must implement.

  \item \texttt{graph.py}: Handles scheduling, execution, and data management throughout the merge process. This is the heart of MergeKit's performance and resource efficiency. Understanding this module is important to ensure that intermediate results and data movement across devices is handled efficiently.
  
  \item \texttt{plan.py}: Responsible for creating the computational graph for a merge. If a new merging strategy has different steps involved or inputs required in combining multiple models, they should be accommodated here.

  \item \texttt{architecture.py}: This module deals with the structures of different checkpoints. Most model architectures are defined using simple JSON files. To add support for odd or unique architectures you may need to modify this file.
\end{itemize}




\begin{table*}[!ht]
\centering
\tiny
\begin{tabular}{|l|c|c|c|c|c|c|}
\hline
\textbf{Model} & \multicolumn{3}{c|}{\textbf{Medical Benchmarks}} & \multicolumn{3}{c|}{\textbf{General Benchmarks}} \\ \cline{2-7} 
 & \textbf{USMLE} & \textbf{MedMCQA} & \textbf{PubMedQA} & \textbf{Arc Challenge} & \textbf{HellaSwag} & \textbf{MMLU} \\ \hline
Llama2-7B-Chat \cite{touvron2023llama} & 35.90 & 35.45 & 73.40 & 44.20 & 55.40 & 46.37 \\ \hline
Meditron-7B \cite{chen2023meditron} & 38.40 & 24.07 & 71.40 & 40.20 & 54.50 & 33.06 \\ \hline
MeditronLlama-7B-Lerp & 39.10 & 36.65 & \textbf{75.60} & 46.76 & 58.66 & \textbf{48.44} \\ 
\hline
MeditronLlama-7B-Slerp & \textbf{39.20} & \textbf{36.91} & \textbf{75.60} & \textbf{46.84} & \textbf{58.67} & 47.97 \\ \hline
MeditronLlama-7B-Dare-Ties & 36.37 & 27.56 & 72.20 & 42.92 & 54.79 & 41.17 \\ \hline
MeditronLlama-7B-Ties & 38.73 & 32.27 & \textbf{75.60} & 45.05 & 58.23 & 45.03 \\ \hline
\end{tabular}
\caption{
Comparison of the Llama2-7B Chat and Meditron-7B~\cite{chen2023meditron} models, plus their merged variants, using MergeKit techniques across medical and general benchmarks.  It highlights the best-performing models in bold for each metric.}
\label{tab:medical_model_merging}
\end{table*}

\subsection{Practical Example: Applying Model Merging in Medical Domain}
As illustrated in Table \ref{tab:medical_model_merging}, we experimented with a range of merging techniques available in MergeKit, including Linear intERPolation (LERP), SLERP, TIES, and DARE-TIES, to merge Meditron-7B\footnote{Meditron-7B checkpoint is based on Llama2-7B base model, which is extensively pretrained on a comprehensively curated medical corpus.} \cite{chen2023meditron} with the Llama2-7B chat model \cite{touvron2023llama}. Both models are based on the Llama2-7B base model. The evaluation results are depicted in Table \ref{tab:medical_model_merging}. According to the findings, all the merged models outperform the Meditron-7B model across various medical benchmarks, including the US Medical
License Exam (USMLE) \cite{jin2021disease}, Medical Multiple-Choice Question Answering (MedMCQA) \cite{pal2022medmcqa}, and PubMed\footnote{\href{https://pubmed.ncbi.nlm.nih.gov/}{https://pubmed.ncbi.nlm.nih.gov/}} Question Answering (PubMedQA) \cite{jin-etal-2019-pubmedqa}. Furthermore, models merged using LERP and SLERP techniques exhibit superior performance over the Llama2-7B chat model in general benchmarks. Our empirical experiments highlight the varying capabilities of merged models and provide comparative performance insights. Within the medical domain, the SLERP method appears to outperform others. However, more importantly, these experiments reveal how model merging can lead to the development of more generalized models with enhanced capabilities across diverse applications.

Recent studies emphasize the importance of merging fine-tuned models into their base models to address challenges like catastrophic forgetting and skill transfer~\cite{alexandrov2024mitigating, siriwardhana2024domain}. This technique helps maintain prior knowledge while integrating new capabilities. We employed several merging techniques, each with its own hyper-parameters, such as the contribution of each pre-trained model and parameter masking in task vectors.

\section{Conclusion and Future Work}

In this paper, we introduce MergeKit, an innovative open-source tool for seamlessly integrating LLMs. We detail its functionalities and provide an overview of recent model merging literature from an engineering perspective. Additionally, we offer insights on incorporating new merging techniques, encouraging community contributions. MergeKit is a dynamic project, committed to continuously integrating new methodologies through collaborative efforts with the open-source community.


\section*{Ethical Considerations}
As stewards of the open-source community dedicated to the advancement of LLMs, our work with
MergeKit underscores a commitment to democratizing access to cutting-edge AI technologies while
fostering an environment of ethical integrity and
continuous improvement. By providing an open-source toolkit that enables the merging of model
checkpoints, we aim to enhance the collaborative
capabilities of researchers, developers, and practitioners across the globe, encouraging innovation
and the sharing of knowledge. In doing so, we
are acutely aware of the necessity to uphold principles of fairness, accountability, and transparency
within this community. This includes the proactive identification and mitigation of biases within
merged models, ensuring the ethical use of data,
and maintaining the privacy and security of information. Our commitment extends beyond technological advancements, encompassing the responsibility to engage with diverse stakeholders, gather
feedback, and adapt our approaches to address ethical concerns effectively. We recognize the imperative to continually evolve our practices, striving for
solutions that not only push the boundaries of AI
but also do so with an unwavering commitment to
the improvement of society.

\bibliography{anthology,custom}
\bibliographystyle{acl_natbib}




\end{document}